\documentclass[sigconf]{acmart}

\AtBeginDocument{%
  \providecommand\BibTeX{{%
    \normalfont B\kern-0.5em{\scshape i\kern-0.25em b}\kern-0.8em\TeX}}}

\copyrightyear{2023}
\acmYear{2023}
\setcopyright{acmlicensed}\acmConference[WWW '23]{Proceedings of the ACM
Web Conference 2023}{April 30-May 4, 2023}{Austin, TX, USA}
\acmBooktitle{Proceedings of the ACM Web Conference 2023 (WWW '23), April
30-May 4, 2023, Austin, TX, USA}
\acmPrice{15.00}
\acmDOI{10.1145/3543507.3583368}
\acmISBN{978-1-4503-9416-1/23/04}

\usepackage[T1]{fontenc}
\usepackage{enumitem}
\usepackage{times}
\usepackage{soul}
\usepackage{url}
\usepackage{hyperref}
\usepackage{inputenc}
\usepackage{caption}
\usepackage{graphicx}
\usepackage{amsmath}
\usepackage{amsthm}
\usepackage{booktabs}
\usepackage{algorithm}
\usepackage{algorithmic}
\usepackage{multirow}
\usepackage{amsfonts}
\newtheorem{prop}{Proposition}

\begin{document}

\title{Graph Neural Network with Two Uplift Estimators for Label-Scarcity Individual Uplift Modeling}

\author{Dingyuan Zhu$^{1*}$, Daixin Wang$^{1*}$, Zhiqiang Zhang$^1$, Kun Kuang$^{2}$, \\
Yan Zhang$^1$, Yulin Kang$^1$, Jun Zhou$^{1\#}$}
\affiliation{
  $^1$Ant Group, China \\
  $^2$Zhejiang University
  \country{China}
}
\email{
  {dingyuan.zdy,daixin.wdx,lingyao.zzq,zy118409,yulin.kyl,jun.zhoujun}@antgroup.com, kunkuang@zju.edu.cn
}
\thanks{* Equal contribution. \\
\# Corresponding author.}

\begin{abstract}
Uplift modeling aims to measure the incremental effect, which we call uplift, of a strategy or action on the users from randomized experiments or observational data. Most existing uplift methods only use individual data, which are usually not informative enough to capture the unobserved and complex hidden factors regarding the uplift. Furthermore, uplift modeling scenario usually has scarce labeled data, especially for the treatment group, which also poses a great challenge for model training. Considering that the neighbors' features and the social relationships are very informative to characterize a user's uplift, we propose a graph neural network-based framework with two uplift estimators, called GNUM, to learn from the social graph for uplift estimation. Specifically, we design the first estimator based on a class-transformed target. The estimator is general for all types of outcomes, and is able to comprehensively model the treatment and control group data together to approach the uplift. When the outcome is discrete, we further design the other uplift estimator based on our defined partial labels, which is able to utilize more labeled data from both the treatment and control groups, to further alleviate the label scarcity problem. Comprehensive experiments on a public dataset and two industrial datasets show a superior performance of our proposed framework over state-of-the-art methods under various evaluation metrics. The proposed algorithms have been deployed online to serve real-world uplift estimation scenarios. 
\end{abstract}


\maketitle

\section{Introduction}
Uplift modeling refers to the set of techniques used to estimate the effect of an action on a user's outcome. This technology can be applied to various fields, such as economics\cite{olaya2020survey,gubela2017revenue}, medicine\cite{jaroszewicz2014uplift,jaskowski2012uplift} and sociology\cite{olaya2020uplift,guo2020counterfactual}. For example, an e-commerce company is preparing to send promotional coupons (i.e., action or treatment) to some users to attract them to purchase more products (i.e., outcome). Then estimating the individual uplift can help to find target users. Uplift modeling is a complicated problem because one needs to estimate the difference between two outcomes with and without the treatment which are mutually exclusive to individuals. For example, we can only observe the outcome of a user getting or not getting a promotional coupon. The outcome we can observe is called the factual outcome and the outcome we can not observe is called the counterfactual outcome. Therefore, uplift modeling also can be seen as a counterfactual inference problem\cite{gutierrez2017causal,kuang2020data}. 

To model the counterfactual outcome, existing uplift methods mainly rely on randomized experiments or observational data: users will be assigned to either the treatment group or the control group and we can observe only one type of outcome. Based on these data, most existing methods \cite{gutierrez2017causal,guelman2015uplift,rzepakowski2012decision} utilize user's individual features to estimate the user uplift \cite{breiman2001random,chen2016xgboost}. An accurate uplift modeling is required to capture the complex and the unobserved hidden factors (representations) within the user. However, many factors regarding the user's uplift are difficult to be captured by only using individual data for two reasons. Firstly, in real-world scenarios, the individual features of users, especially the new users, are missing. Secondly, some informative information for uplift estimation is hidden and difficult to be characterized by individual features, like social status and personality.

To resolve the aforementioned problems, we hope to introduce the social graph. On the one hand, considering that users with close social relations usually have similar behaviors and preferences, we can utilize the information from social neighbors as a supplement to the user's own features. On the other hand, social graphs can reveal some informative social information like social status, which is also beneficial for uplift modeling \cite{veitch2019using}. From the experiments, we also demonstrate that the uplift difference between users with friend relationships is much smaller than the difference between random users. Therefore, it is very essential to incorporate social graphs when doing uplift estimation. 

A few of the existing works propose graph-based uplift models for uplift estimation \cite{veitch2019using,guo2020counterfactual,guo2020learning}: They first use graph neural network (GNN) model to learn the graph-based representations from the social graphs. Then with the graph-based representations, these works design uplift estimators, which consist of two separate pathways of models to predict the outcome with and without the treatment using the data of the treatment group and control group respectively. Although these works have achieved substantial improvements, they still have limitations. In real-world scenarios, since the effect of the treatment is unpredictable, imposing the treatment on samples may bring a bad effect like capital loss and user loss. In this way, it is common that only a little flow will be assigned to the treatment group. In this way,  the number of labeled instances, particularly the instances from the treatment group, is commonly limited, especially in the scenario of randomized experiments we mainly focus on. Moreover, to model the relational data, graph-based models require learning more parameters. From Figure \ref{sec:scarcity}, compared with tree-based and NN-based models, the performance of existing graph-based models will drop more quickly when the labeled data is scarce, thereby labeled data scarcity is a more severe and unresolved problem for graph-based representation learning in uplift estimation. 

To address aforementioned problems, we propose a general GNN-based framework with two uplift estimators for uplift modeling. First of all, a GNN-based model with breadth and depth aggregators is proposed to generate the graph-based representations for the following uplift estimation. And empirically we demonstrate the proposed uplift estimation framework is general for many existing GNN-based models. Furthermore, to address the label scarcity problem, we propose two uplift estimators to utilize the treatment and control group data and their corresponding social relations. Firstly, we design a class-transformed target, which we prove is equal to the uplift and is general for uplift scenarios. Unlike two separate pathways of estimators, our transformed target is able to utilize the training instances with and without the treatment simultaneously in a common model to approach the uplift. Furthermore, when the outcome is discrete, we design partial labels based on the user's treatment and the observed outcome. Then we design the other uplift estimator with two classifiers to learn partial labels. The two classifiers can focus on different facets of the uplift but all require two groups of data for training. In this way, our model can capture the relations between two groups of data and utilize the labeled data more effectively. Experimental results show our proposed framework can outperform the best-performing baseline method by an improvement of $5$\% to $10$\% in the regression setting and $12$\% to $25$\% in the classification setting. 

The main contributions of the paper are summarized here:
\begin{itemize}
    \item \textbf{Problem}: We point out the label scarcity problem is severe for uplift modeling, especially when trained with the graph-based model. To the best of our knowledge, it is the first work trying to solve the label scarcity problem in uplift estimation.
    \item \textbf{Methodology}: We introduce a novel GNN-based framework with two uplift estimators for uplift modeling, which can utilize the two groups of data and the social graph comprehensively. Specifically, when the outcome is discrete, we are the first to introduce partial label learning to uplift estimation, which is able to utilize the labeled data more effectively to alleviate the labeled data scarcity problem.
    \item \textbf{Results}: Extensive experiments on a public dataset and two industrial datasets demonstrate the superiority of proposed GNN-based uplift model(GNUM) on different types of outcomes. Specifically, we find labeled data scarcity is indeed a serious problem for previous graph-based uplift methods and our methods are robust to the label scarcity problem. 
\end{itemize}

\section{Related Work}
\subsection{Uplift modeling} \label{sec:uplift_model}
Existing uplift methods can be classified into three categories \cite{gutierrez2017causal}, i.e. the Two-Model methods \cite{radcliffe2007using,nassif2013uplift}, the Class-Transformation methods \cite{jaskowski2012uplift} and the methods that model uplift directly \cite{radcliffe2011real,rzepakowski2012decision,soltys2015ensemble}. 

The Two-Model methods construct two independent models for the two groups of data. One model infers the label using the data from the treatment group and the other model is for the control group data. However, \cite{radcliffe2011real} points out that the Two-Model methods may miss the uplift signal. Then The Class Transformation methods are introduced by \cite{jaskowski2012uplift,athey2015machine}, which aim to create a new target to approach the uplift. Then a single model is proposed to learn the new target. But Class Transformation methods usually require a balanced dataset between the control and treatment groups. The last type of uplift method aims to directly infer the uplift. The work \cite{zaniewicz2013support} proposes a method based on a modification of the SVM model and the work \cite{guelman2014optimal} focuses on k-nearest neighbors to do the uplift estimation. 

The aforementioned works assume that the data of the control group and treatment group are randomly collected. If the collecting data is naturally observed, besides uplift estimation, uplift modeling also needs to reduce the bias of the data from the treatment group and control group. The most popular methods of this type are the doubly robust learning methods \cite{emsley2008implementing,zhao2015doubly,jung2021estimating,chernozhukov2018double}. They usually adopt the Inverse Propensity Scoring (IPS) \cite{hirano2003efficient,kitagawa2018should} to re-weight each instance, aiming at making the uplift estimator unbiased. And some methods\cite{louizos2017causal,shalit2017estimating} which estimate individual treatment effect can be used to estimate the uplift.

However, aforementioned uplift methods assume that the uplift can be fully estimated by the individual features. As we have stated before, the social relationships between users are important for uplift estimation. Then Guo et. al.\cite{guo2020learning,guo2020counterfactual,ma2021deconfounding} first introduce the networked observational data to the problem of causal effects estimation. NetEst\cite{jiang2022estimating} formalize the networked causal effects estimation to a multi-task learning problem and HyperSCI\cite{ma2022learning} learning causal effects on hypergraphs. These methods prove that networked data is important for predicting causal effects. Although incorporating the graph data will violate the Stable Unit Treatment Value Assumption(SUTVA), following works \cite{sobel2006what} have pointed out that SUTVA is not plausible in real-world scenarios. We will follow these works \cite{guo2020learning,guo2020counterfactual} to incorporate the graph data but are distinct from them by addressing the label scarcity problem for graph-based uplift estimation.  

\subsection{Partial Label Learning}
Partial label learning deals with the problem that each sample is associated with a set of candidate labels, among which only one label is the ground-truth label to be predicted. Existing partial label learning methods can be classified into two categories: the averaged-based strategy and the identification-based strategy. The averaged-based strategy assumes that each candidate label contributes equally to the model training \cite{hullermeier2006learning,gong2017regularization}, which may suffer from the problem that the real label is overwhelmed by other labels. To overcome this drawback, the identification-based methods give different confidence to different candidate labels by learning the topological information \cite{zhang2015solving,zhang2016partial}. Existing partial-label learning is often applied to automatic face naming, object detection, and web mining. As far as we know, this is the first work to apply partial label learning to uplift modeling.

Other typical types of weakly-supervised learning  include incomplete label learning\cite{zhao2015semi-supervised} and inaccurate label learning\cite{wu2018a}. They are not closely related to our problem, which will not be discussed here. 

\subsection{Graph Neural Network}
Graph neural networks (GNNs), aiming to generalize neural networks to deal with graph data, have drawn increasing research interest recently \cite{bronstein2017geometric,zhang2020deep} and show effectiveness in various tasks \cite{wang2017community,fan2020graph,zhang2021graph}. Generally, current GNNs can be divided into two categories: spectral-based methods and spatial-based methods. Spectral-based GNNs are originated from signal processing and are commonly based on the Laplacian Matrix \cite{bruna2014spectral,defferrard2016convolutional,kipf2017semi}.  Spatial-based GNNs regard the graph convolution as the 'message-passing' framework in the spatial domain, i.e. defining the graph convolution as nodes aggregating information from neighborhoods  \cite{gilmer2017neural,hamilton2017inductive}. And \cite{feng2021should, sui2022causal, ding2022causal} have explored the causal inference problems with GNN-based models. More GNN-based models can be referred in recent surveys \cite{wu2020comprehensive,zhang2020deep}. However, only a few works focus on uplift estimation using GNNs and they do not address some specific and important problems when using GNNs on uplift estimation.

\section{Model Formulation}

\subsection{Notations and Preliminaries}
Firstly, we describe the notations used in this paper. We denote a scalar with a letter (e.g., $t$), a vector with a boldface lowercase letter (e.g., $\mathbf{x}$), and a matrix with a boldface uppercase letter (e.g., $\mathbf{A}$).

In our uplift estimation problem, we assume there are $N$ users in total. The data of each user $i$ can be represented as $\{\mathbf{x}_i,t_i,y_i\}$, where $\mathbf{x}_i \in \mathbb{R}^d$ represents the individual feature vector, $t_i\in\{0,1\}$ denotes the observed treatment and $y_i$ denotes the observed outcome. Note that $y_i(1)$ denotes the outcome of user $i$ when he receives the active treatment, and $y_i(0)$ denotes its outcome  with the control treatment. In this paper, we focus on the scenario of a randomized experiment, which means that each user is randomly given the treatment or not. Then, the actual uplift of user $i$ is defined as: 
\begin{equation} \label{eq_up}
	\tau_i = y_i(1) - y_i(0),
\end{equation}
and our target in this paper is to estimate the uplift of each user. 

However, for a specific user $i$, we can only observe $y_i(1)$ or $y_i(0)$. The one we can observe is called the factual outcome. And the other one we cannot observe is called the counterfactual outcome. It is not difficult to find that the key and the challenging issue of uplift modeling is to do the counterfactual prediction. As we have stated before, the user's social relationships and his/her social neighbors' features contribute a lot to uplift estimation. Therefore, we introduce the social graph in our work. 

We define the social graph as $\mathit{G}=(\mathit{V},\mathit{E})$ \footnote{For simplicity, we assume the social graph is a directed unweight graph}, where $\mathit{V}=\{v_1,...,v_N\}$ denotes the set of nodes, $N = \left| \mathit{V} \right|$ is the number of nodes, and $\mathit{E} \subseteq \mathit{V} \times \mathit{V}$ is the set of edges between nodes. Here the node denotes a user and the edge denotes two users' social relationships.  Let $\mathbf{X}$ be a matrix of node attributes.
We define $\mathbf{H}^{(l)} = \left[ \mathbf{h}^{(l)}_1,\mathbf{h}^{(l)}_2,...,\mathbf{h}^{(l)}_N \right]$ as the hidden representations of nodes in the $l^{th}$ layer of the graph neural networks where $\mathbf{h}^{(l)}_i$ is the representation of node $v_i$. And we use $L$ as the number of layers for the GNN model. For convenience, we also denote $\mathbf{X}$ as $\mathbf{H}^{(0)}$. 

\begin{figure*}[htb]
	\centering
	\includegraphics[width=2.0\columnwidth]{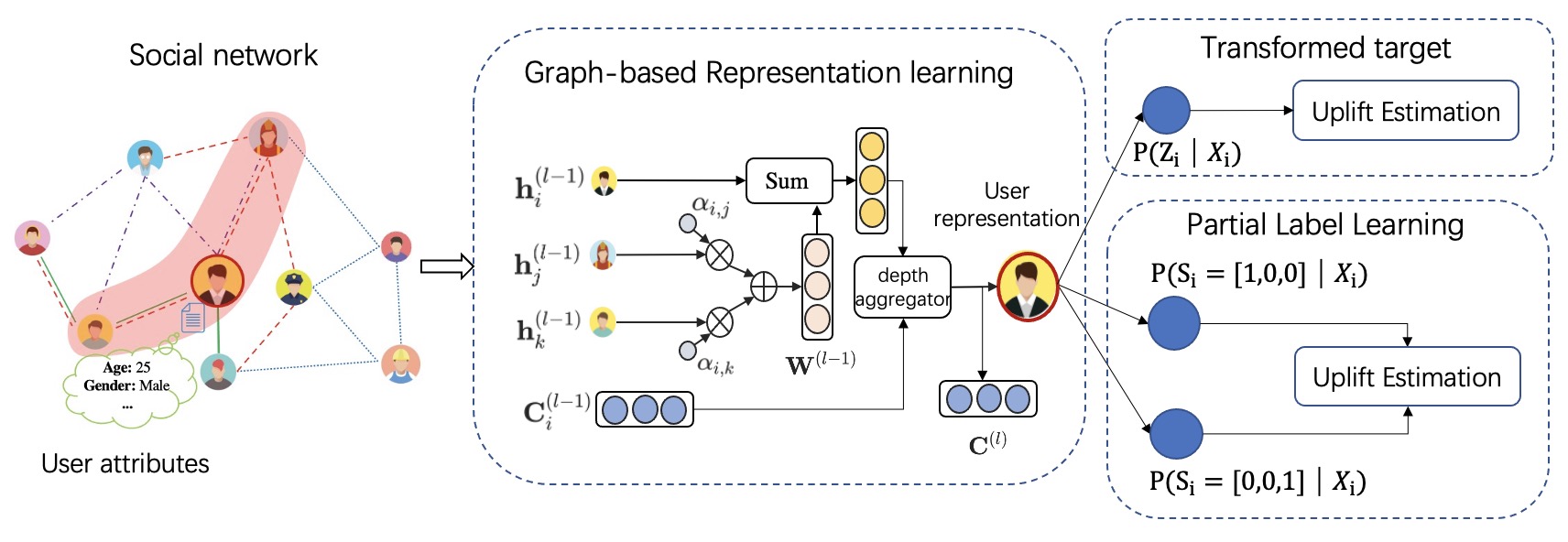} %
	\caption{Framework of GNUM}
	\label{fig:model}
\end{figure*}

\subsection{GNUM}
To incorporate the social relations and the neighbors' attributes, we propose a novel GNN-based uplift model (GNUM). Figure \ref{fig:model} shows the overall framework of the proposed GNUM, which consists of two components, i.e. graph-based representation learning and uplift estimation. Specifically, we propose two GNN-based uplift estimators in this paper working for different scenarios to address the labeled data scarcity problem for uplift estimation scenarios. 

\subsubsection{Graph-based Representation Learning}
This component aims to learn the graph-based node (user) representations by extracting the key information from the social relations and neighbors' attributes. Basically, most GNNs can be represented by: 
\begin{equation} \label{eq_hdl}
	\begin{gathered}
		 \mathbf{\tilde{h}}^{(l+1)}_i = AGGR(\mathbf{\tilde{h}}^{(l)}_i, \mathbf{\tilde{h}}^{(l)}_{\mathcal{N}_i}),
	\end{gathered}
\end{equation}
where $\mathcal{N}_i$ represents the neighbors of user $i$ and AGGR represents the aggregation function of the target user's embedding and his neighbors' embeddings. Experimentally, we demonstrate our proposed uplift estimation framework works well with different aggregators like GCN and GAT. 

Inspired by \cite{liu2019geniepath}, we propose more comprehensive graph-based aggregators in GNUM, which consist of a breadth aggregator to learn information from social neighbors of the current layer and a depth aggregator to ensemble information from different layers. The $l$-th graph convolution layer of GNUM is defined as:
\begin{equation} \label{eq_hdli}
	\begin{gathered}
		 \mathbf{\tilde{h}}^{(l+1)}_i = tanh \left(\sum_{j \in ne(i) \cup \{i\}} 
		\boldsymbol{\alpha}(\mathbf{h}_i^{(l)},\mathbf{h}_j^{(l)})
		 \mathbf{h}^{(l)}_j  \mathbf{W}^{(l)} \right).
	\end{gathered}
\end{equation}
where $\boldsymbol{\alpha}(\mathbf{h}_i^{(l)},\mathbf{h}_j^{(l)})$ is the attention function of the breadth aggregator to measure the importance of user $i$ and user $j$ for uplift modeling, defined as:
\begin{equation} \label{eq_att}
	\begin{gathered}
		\boldsymbol{\alpha}(\mathbf{h}_i^{(l)},\mathbf{h}_j^{(l)}) = softmax(\mathbf{v}^{(l)}tanh(\mathbf{W}_{s}\mathbf{h}_i^{(l)} + \mathbf{W}_{d}\mathbf{h}_j^{(l)})),
	\end{gathered}
\end{equation}
where  ${\mathbf{W}_s}^{(l)}$ represents the weight for the source node, ${\mathbf{W}_d}^{(l)}$ represents the weight for the target node and $\mathbf{v}^{(l)}$ denotes a vector to map the representations to a value.

Given a user $u_i$, the breadth aggregator in each layer will adaptively gather the information from his neighbors and his own representations of the previous layer. Then we further stack multiple convolution layers and utilize a memory-based depth aggregator to aggregate the user embedding, defined as:

$
\begin{array}{ll}
i_{i}=\sigma\left(W_{i}^{(l)^{\top}}  \mathbf{\tilde{h}}_i^{(l+1)}\right), & f_{i}=\sigma\left(W_{f}^{(l)^{\top}}  \mathbf{\tilde{h}}_i^{(l+1)}\right) \\
o_{i}=\sigma\left(W_{o}^{(l)^{\top}}  \mathbf{\tilde{h}}_i^{(l+1)}\right), & \tilde{C}=\tanh \left(W_{c}^{(l)^{\top}}  \mathbf{\tilde{h}}_i^{(l+1)}\right) \\
C_{i}^{(l+1)}=f_{i} \odot C_{i}^{(l)}+i_{i} \odot \tilde{C}, & \mathbf{h}_{i}^{(l+1)}=o_{i} \odot \tanh \left(C_{i}^{(l+1)}\right)
\end{array}.
$

In this way, our proposed method can extract both local and global structural information from the social graph to obtain the graph-based representations for each user, which facilitates the learning of the following uplift estimators.

\subsubsection{Transformed Target-based Uplift Estimator}
\label{sec:uplift}
With the graph-based hidden representations, an intuitive way is to build two pathways of models to project the node representations into the label space, aiming to approach the outcome with and without treatment using the treatment group data and control group data respectively \cite{guo2020learning,guo2020counterfactual}. The process can be formulated as $\hat{y}_i(t) = f^{(t)}(\mathbf{h}_i^{(L)}), t\in\{0,1\}$, where $f^{(t)}(\cdot)$ is one pathway of the model for treatment or control group data and $\hat{y}_{i}(t)$ represents the prediction of the user's outcome with the treatment or not. Note that $\hat{y}_{i}(t)$ can be continuous as the regression task or discrete as the multi-classification task. 

By optimizing the regression loss or classification loss, the uplift of user $i$ can be estimated as $\hat{\tau}_i = \hat{y}_{i}(1) - \hat{y}_{i}(0)$. However, any one pathway of their model can only utilize one group of data to learn, which will suffer from the labeled data scarcity problem. Furthermore, as previous works of literature state \cite{gutierrez2017causal}, Two-Model methods cannot well capture the relationship between data of the treatment group and the control group, which limits their performance on uplift modeling.  

To address the problem, we propose a class-transformed target and prove that by using the transformed target as the learning objective for both treatment and control group data, the model is equal to do uplift estimation. We first define the observed outcome of user $i$ as:
\begin{equation}
y_i^{obs} = t_iy_i(1) + (1 - t_i)y_i(0).
\label{eqn:y_obs}
\end{equation}

Then the class-transformed target can be defined as follows:
\begin{equation}  \label{eq_zi}
	z_i = y_i^{obs}\cdot \frac{t_i-p}{p(1-p)},
\end{equation}
where $p=\mathbb{E}(t_i|\mathbf{x}_i,\mathit{G})=\mathbb{E}(t_i)=p(t_i=1)$ is defined as the probability that user $i$ receives the treatment. Since we focus on the randomized experiment, it is a constant. 

We have the following important proposition: 
\begin{prop}
The uplift of user $i$ can be estimated in the following form: $\hat{\tau}_i=\mathbb{E}(z_i|\mathbf{x}_i, \mathit{G})$.
\end{prop}

\renewcommand{\qedsymbol}{}
Inspired by \cite{athey2015machine}, the proof of the proposition can be found in the Appendix \ref{appx:proof}.

Based on the proposition, we can project the node representations $\mathbf{H}^{(L)}$ to form the transformed target. The loss for the GNUM with class-transformed target (GNUM-CT) is:
\begin{equation}\label{eq:loss_ct}
	\mathcal{L}_{CT} =  \sum_{i \in \mathit{V}}{( {z}_{i} -\hat{z}_{i})^2}+\lambda L_{reg},
\end{equation}
where $z_i$ has been given in Eq. \eqref{eq_zi}, $\hat{{z}}_{i}= \sigma(\mathbf{W}_{CT}\mathbf{h}_i^{(L)}+\mathbf{b}_{CT})$ is the prediction of the target, $\mathbf{W}_{CT}$ and $\mathbf{b}_{CT}$ are learnable parameters. $L_{reg}$ is defined as the $L2$ regularized loss on all parameters of the proposed model and $\lambda$ is set to $0.0005$ in this paper.   

In summary, both the treatment group and control group data are utilized to learn and optimize the proposed transformed uplift estimator, which can well alleviate the label scarcity problem and capture the inherent relationship between the two groups of data. Moreover, we do not make any assumption regarding the proposed uplift estimator with the class-transformed target. It can work well for any types of outcome, i.e. continuous and discrete outcome. Additionally, the newly proposed target is also general to balanced and imbalanced data of two groups. If the two groups of data are biased, we can also replace the $p$ in Eq. \eqref{eq_zi} with the sample's propensity score. Therefore, it is a very general uplift estimator which can comprehensively utilize the two groups of data for modeling individual uplift values. 
 
\subsubsection{Partial-Label-based Uplift Estimator}
When the outcome is discrete, i.e. a multi-classification outcome prediction problem,  we propose a partial-label-based uplift estimator to further utilize more labeled data for uplift estimation. For simplicity, we will assume that the outcome is binary and introduce our solution. It can be generalized to a multi-classification problem by transforming the problem into multiple binary scenarios. 

In detail, we divide the whole users into three groups:
\begin{itemize}
	\item Group A: The group of users who give positive outcomes regardless of whether they receive the treatment.
	\item Group B: The group of users who give positive outcomes only when they receive the treatment.
	\item Group C: The group of users who do not give positive outcomes regardless of whether they receive the treatment.
\end{itemize}
Without loss of generality, in this paper, we consider the cases where the treatment has a positive impact on the user's outcome (e.g., promotional coupon). Thus the situation where a user gives a negative outcome with the treatment but gives a positive outcome without the treatment does not exist. 

Next, we generate a 3-bits coding as the partial label for each user. Each bit of code corresponds to whether the user belongs to the above-mentioned group. 
According to the treatment and observed outcome, the partial label of $S_i = [s_{i}^A, s_{i}^B, s_{i}^C]$ for user $i$ is defined as:
\begin{equation}
S_i =
\begin{cases}
[1, 0, 0]& \: if \: t_i=0 \: and \: y_i^{obs}=1\\
[0, 1, 1]& \: if \: t_i=0 \: and \: y_i^{obs}=0\\
[1, 1, 0]& \: if \: t_i=1\: and \: y_i^{obs}=1\\
[0, 0, 1]& \: if \: t_i=1\: and \: y_i^{obs}=0.
\end{cases}
\end{equation}

Taking a user belonging to $t_i=1, y_i^{obs}=1$ as an example: if we give the treatment to the user $i$, we can observe his positive outcome. Then based on our aforementioned definition, the user may belong to Group A or Group B. Therefore, based on our definition, the partial label is $[1, 1, 0]$ for user $i$. 

We build two binary classifiers to help estimate the uplift. The first classifier gives the probability that the user belongs to group A, i.e. $P(S_i=[1, 0, 0]|\mathbf{x}_i)$. Then the instances whose partial labels are $S_i = [1, 0, 0]$ are regarded as positive samples, and the instances with the partial label $S_i \in \{[0, 1, 1], [0, 0, 1]\}$ are negative samples for the first classifier. The second classifier gives the probability that the user belongs to group C, i.e. $P(S_i=[0, 0, 1]|\mathbf{x}_i)$. Thus the instances with the partial label $S_i = [0, 0, 1]$ are positive samples, and the instances with $S_i \in \{[1, 1, 0], [1, 0, 0]\}$ are negative samples. 

Then we project the node representations to the partial label space:
\begin{equation} 
\begin{split}
    \hat{y}_i^{pl1} = \sigma(\mathbf{W}^{pl1}\mathbf{h}_i^{(L)}+\mathbf{b}^{pl1})  \\
    \hat{y}_i^{pl2} = \sigma(\mathbf{W}^{pl2}\mathbf{h}_i^{(L)}+\mathbf{b}^{pl2}), 
\label{eqn:partial_resp}
\end{split}
\end{equation}
where $\hat{y}_i^{pl1}$ and $\hat{y}_i^{pl2}$ represent the prediction of the two partial labels, $\mathbf{W}^{pl1}$, $\mathbf{W}^{pl2}$, $\mathbf{b}^{pl1}$ and $\mathbf{b}^{pl2}$ are learnable parameters.

Then we define the cross-entropy loss as the classification loss for partial labels. The overall loss function for GNUM with partial-label-based estimator (GNUM-PL) is defined as: 
\begin{equation}
	\mathcal{L} =  \sum_{i \in {V}} 
	(\mathcal{F}(y_i^{pl1}, \hat{y}_{i}^{pl1}) + \mathcal{F}(y_i^{pl2}, \hat{y}_{i}^{pl2}))+\lambda  L_{reg}, 
\label{eq:gnum_loss}
\end{equation}
where $y_i^{pl1}$ and $y_i^{pl2}$ are the ground truth for the first and second partial labels we defined. $\mathcal{F}(y_i, \hat{y}_{i})$ is defined as the cross-entropy loss, where $\mathcal{F}(y_i, \hat{y}_{i})=- ( {y}_{i} \log \hat{y}_{i} - (1- {y}_{i}) \log (1 -\hat{y}_{i}))$. $L_{reg}$ is defined as the $L2$ regularized loss on all the parameters of the model and $\lambda$ is set to $0.0005$ in this paper.  

Using the above loss function, the proposed graph model with partial labels can be trained. Then the final uplift of user $i$ can be defined in the following ways:
\begin{equation} 
\begin{split}
	 \hat{\tau}_i &= p(y_i(1)|\mathbf{x}_i,\mathit{G}) - p(y_i(0)|\mathbf{x}_i),\mathit{G}) \\
	              &= 1 - P(S_i=[0, 0, 1]|\mathbf{x}_i,\mathit{G}) - P(S_i=[1, 0, 0]|\mathbf{x}_i,\mathit{G})  \\
	              &= 1-\hat{y}_i^{pl1}-\hat{y}_i^{pl2} \nonumber 
\end{split}.
\end{equation}

Now we introduce why the proposed model with the partial label can further improve the performance. Although the graph-based model with the transformed target we just proposed in Section \ref{sec:uplift} utilizes a general target for both treatment group data and control group data to capture the relationship between two groups of data, each sample can only be utilized once in each epoch of training. However, in our partial-label-based uplift estimator, both classifiers will utilize both the treatment data and the control data but focus on different facets of information. Specifically, the data with $S_i=[1,1,0]$ and $S_i=[0,1,1]$ can be utilized by both classifiers. In this way, on the one hand, more labeled data can be utilized to train each classifier, ensuring a better performance especially when the labeled data is scarce in uplift estimation scenarios. On the other hand, since the two classifiers can both obtain the two groups of data, the classifiers can better capture the relations and learn useful information from the two groups to achieve better results. 

The pseudo-code and complexity analysis of GNUM-CT and GNUM-PL can be found in Appendix \ref{appx:code}.

\section{Experiments}
\begin{table*}[h]
\caption{The experimental result on public dataset BlogCatalog. The result of the best performance is in bold and the result of the second best performance is underlined. }
\centering
	\label{tab:ate}  
	\begin{center}	
		\small
		\resizebox{\textwidth}{!}{
			\begin{tabular}{ccccccccccccc}
				\toprule[1pt]
				{$\kappa_2$}&{Outcome Type}&{Metric}&GNUM-PL&GNUM-GCN&GNUM-GAT&GNUM-CT&Two-Model&CTM&Uplift-RF&NetDeconf&DML&DRL\\
				\hline
				\multirow{4}*\textbf{0.5}
                & \multirow{2}*\textbf{Continuous} & {$\sqrt{\epsilon_{PEHE}}$}  &   / & /  & / & \textbf{4.164} &  9.215   & 8.448 & 6.760 & \underline{4.496} & 5.312 & 5.407  \\
				  ~ & ~ & $\epsilon_{ATE}$  &   / & /  & / & \textbf{0.935} &  4.172  & 3.317 & 2.629 & \underline{0.970} & 1.244 & 1.360   \\
                    ~ & \multirow{2}*\textbf{Binary} & {$\sqrt{\epsilon_{PEHE}}$}  &  \textbf{0.529} & 0.642  & \underline{0.586} & 0.613 &  1.253  & 0.964 & 0.740 & 0.621 & 0.689 & 0.707  
                    \\
				~ & ~ & $\epsilon_{ATE}$  &  \textbf{0.121} & 0.147  & \underline{0.129} & 0.138 &  0.470  & 0.364 & 0.210 & 0.135 & 0.159 & 0.172   
                 \\
                \multirow{4}*\textbf{2}
                & \multirow{2}*\textbf{Continuous} & {$\sqrt{\epsilon_{PEHE}}$}  &   / & /  & / & \textbf{9.337} &  23.348   & 17.277 & 15.945 & \underline{9.623} & 12.134 & 13.060  \\
				  ~ & ~ & $\epsilon_{ATE}$  &   / & /  & / & \textbf{2.102} &  10.920  & 8.624 & 8.004 & \underline{2.243} & 6.530 & 7.191  \\
                    ~ & \multirow{2}*\textbf{Binary} & {$\sqrt{\epsilon_{PEHE}}$}  &  \textbf{0.353} & 0.430  & \underline{0.392} & 0.411 &  1.047  & 0.820 & 0.699 & 0.423 & 0.554 & 0.598  
                    \\
				~ & ~ & $\epsilon_{ATE}$  &  \textbf{0.105} & 0.136  & \underline{0.114} & 0.126 &  0.417  & 0.255 & 0.202 & 0.131 & 0.164 & 0.160
                 \\
				\bottomrule[1pt]
		\end{tabular}}
	\end{center}
\end{table*}

In this section, we conduct experimental results to answer the following three questions:
\begin{itemize}
    \item \textbf{Q1}: How our method performs compared with all the baseline methods on a public dataset and two industrial datasets? (Answered in Section  \ref{sec:blogcatalog} and Section \ref{sec:uplift_curve}.)
    \item \textbf{Q2}: How can our proposed estimators generalize to other GNNs? (Answered in Section \ref{sec:gene_GNN}.)
    \item \textbf{Q3}: What is the relationship between the user uplift and the social relation? (Answered in Section \ref{sec:neigh}.)
    \item \textbf{Q4}: How do our proposed graph-based methods and comparing methods perform with different amounts of labeled data? (Answered in Section \ref{sec:scarcity}.)
\end{itemize}

\subsection{Experiment settings}
\subsubsection{Dataset} 
We evaluate our method on one public dataset and two real-world industrial datasets. 

Firstly, we follow \cite{guo2020learning} to  build a semi-synthetic dataset base on BlogCatalog: the node features and network structures are collected from the BlogCatalog. The treatments and outcomes are synthesized. There are confounders in this dataset and the control and test datasets are biased. In detail, the node represents a blogger and the edge denotes their social relationships. The node features are bag-of-words representations of keywords to describe the bloggers. We synthesize (1) the outcomes as the rating of readers on the bloggers and (2) the treatments as whether the blog contents are shown on mobile devices or desktops. The detailed synthetic process for the synthetic process can be found in Appendix \ref{appn:dataset}. There are three different parameters $C, \kappa_1, \kappa_2$ that control the synthetic results of the dataset. Following the experimental settings in previous work\cite{guo2020learning}, we set $C=5, \kappa_1=10, \kappa_2 \in\{0.5,2\}$.

The two real-world industrial datasets are collected from an internet company\footnote{The data set does not contain any Personal Identifiable Information. The data set is desensitized and encrypted. Adequate data protection was carried out during the experiment to prevent the risk of data copy leakage, and the data set was destroyed after the experiment. The data set is only used for academic research, it does not represent any real business situation.}, denoted as Industry-A and Industry-B. The company has two products and plans to send discount coupons to users who have not purchased the products. 
Because the discount coupons are limited,  we need to find the users who are most likely to purchase the product when receiving the coupons.
This can be regarded as a problem of uplift modeling.
For each dataset, the users are randomly split into the treatment group and the control group. The users of the treatment group will receive the coupons and the users of the control group will not. Then we observe whether they will purchase the product for the following $30$ days. The nodes represent users and the edges represent users' friendships. The user features mainly consist of statistical features regarding the user's profile and behaviors in our platform. Note that the collection of data removes the user's sensitive information and obtains the user's privacy authorization.

The detail statistics of three datasets are shown in Table \ref{tab:stat}.

\begin{table}
    \small 
	\begin{center}
		\caption{Statistics of datasets. $|V|$ denotes the number of users, $|E|$ denotes the number of edges and $|F|$ denotes the number of attributes. ${V_t}$ and ${V_c}$ represent the set of users in the treatment group and the set of users in the control group, respectively. }
		\label{tab:stat}
		\small
		\resizebox{0.5\textwidth}{!}{
			\begin{tabular}{cccccc}
				\toprule[1pt]
				\textbf{Datasets}&$|V|$&$|E|$&$|F|$&$|V_t|$&$|V_c|$\\
				\hline
				\textbf{Industry-A}  &  505.2K  &  2.2M  & 652   &252.6K & 252.6K \\
				\textbf{Industry-B}  &  573.8K  &  2.6M  & 915   &286.9K & 286.9K \\
                    \textbf{Blogcatalog}  &  5.2K  &  173.5K  & 8189   & 1.6K & 3.6K \\
				\bottomrule[1pt]
		\end{tabular}}
	\end{center}
	\label{tabdatasets}
\end{table}

\subsubsection{Baseline Methods}
To evaluate the effectiveness of GNUM-PL and GNUM-CT, we compare them with three lines of state-of-the-art uplift methods, including NN-based methods (Two-Model~\cite{radcliffe2007using} and CTM~\cite{jaskowski2012uplift}), a tree-based method (Uplift-RF~\cite{guelman2015uplift}), Causal-effect-based methods (DML~\cite{chernozhukov2018double} and DRL~\cite{zhao2015doubly}) and graph-based uplift methods (NetDeconf~\cite{guo2020learning}). GNUM-GCN and GNUM-GAT are implemented by using the GCN or GAN as the GNN backbones but still use partial-label-based uplift estimator. More details of the comparing methods can be found in Appendix \ref{appn:baseline}. And the parameter settings can be found in Appendix \ref{appn:para}.

\begin{figure*}[htb]
	\centering
	\begin{minipage}{0.24\textwidth}  
		\small
		\centerline{\includegraphics[width=1\columnwidth]{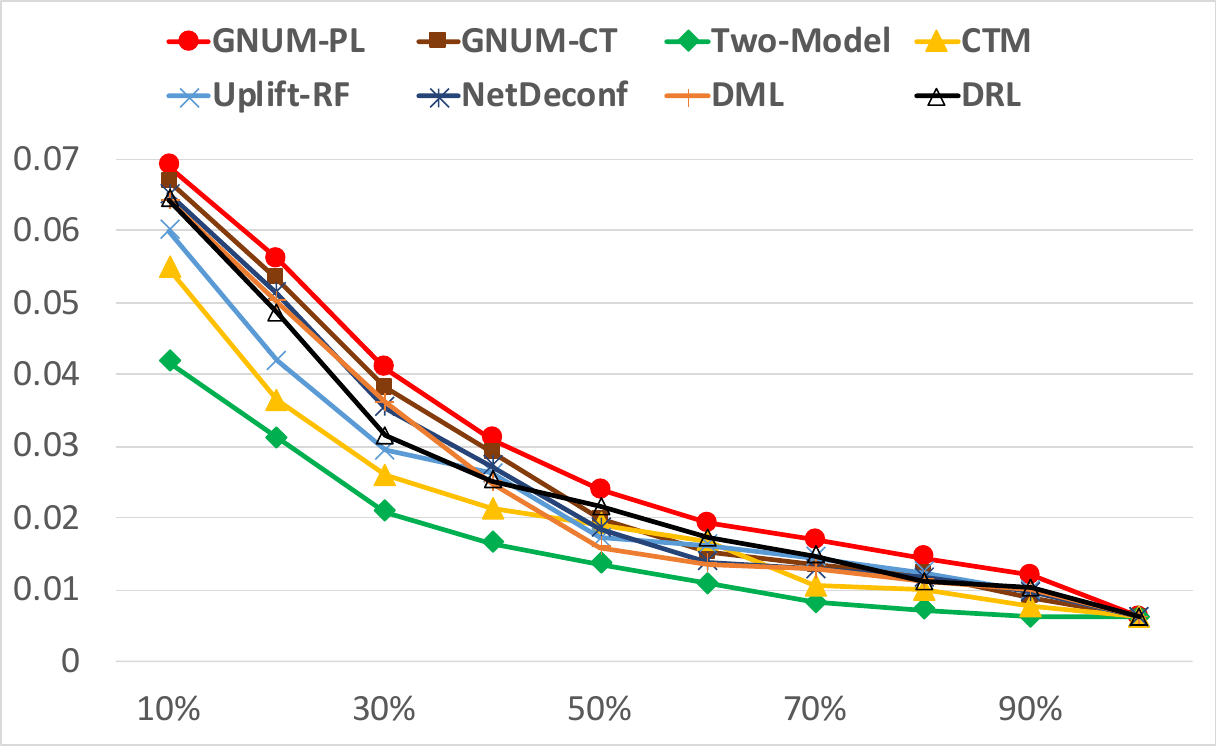}}
		\centerline{(a) Uplift Curve on Industry-A.}
	\end{minipage}
	\begin{minipage}{0.24\textwidth}
		\small
		\centerline{\includegraphics[width=1\columnwidth]{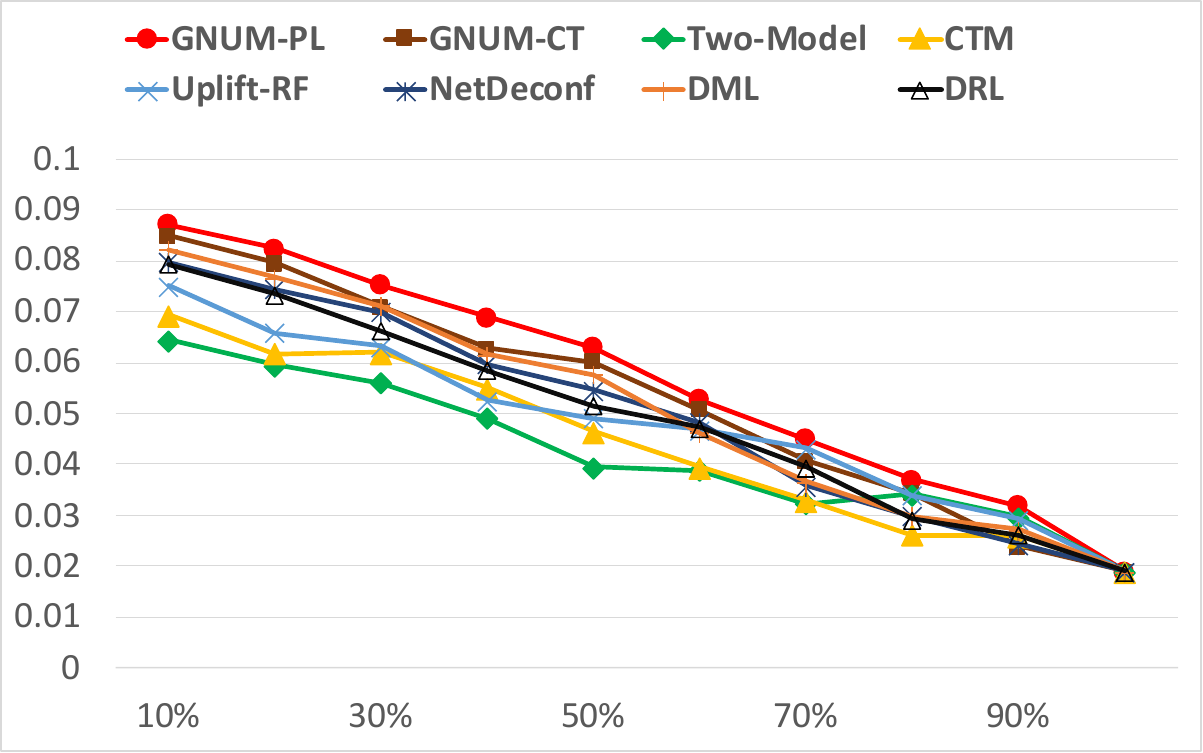}}
		\centerline{(b) Uplift Curve on Industry-B.}
	\end{minipage}
	\begin{minipage}{0.24\textwidth}  
		\small
		\centerline{\includegraphics[width=1\columnwidth]{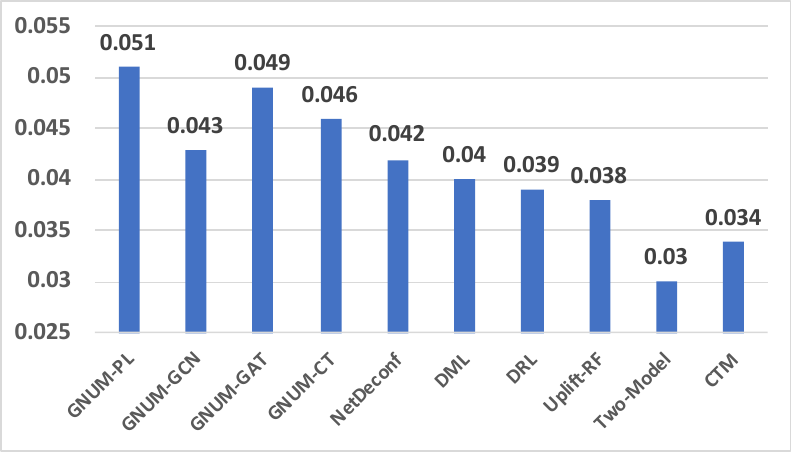}}
		\centerline{(c) Qini Coefficient on Industry-A.}
	\end{minipage}
	\begin{minipage}{0.24\textwidth}
		\small
		\centerline{\includegraphics[width=1\columnwidth]{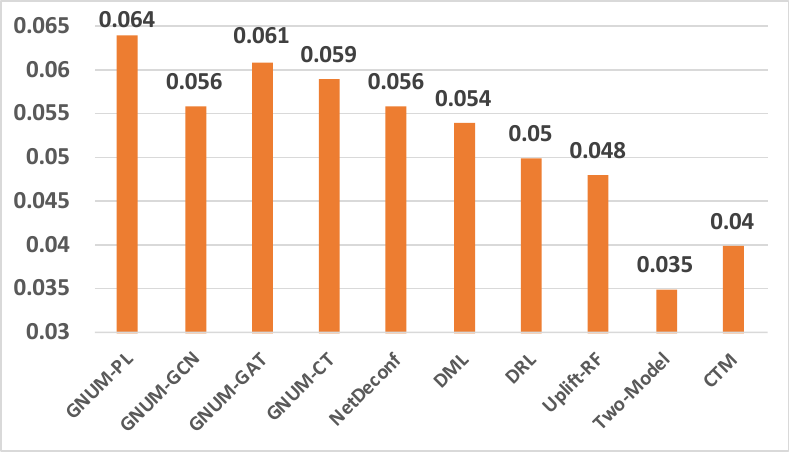}}
		\centerline{(d) Qini Coefficient on Industry-B.}
	\end{minipage}
	\caption{The uplift curve on (a) Industry-A and (b) Industry-B: The ordinate represents the uplift value defined in Eq. \ref{eqn:exp_uplift}, and the abscissa represents users with top $k$\% largest predicted uplift values. The Qini Coefficient on (c) Industry-A and (d) Industry-B.}
	\label{fig:cum_uplift}	
\end{figure*}

\begin{table*}[h]
	\caption{The detailed comparison of the uplift curve. The result of the best performance is in bold and the result of the second best performance is underlined. }
	\centering
	\label{tab:up}  
	\begin{center}	
		\small
		\resizebox{\textwidth}{!}{
			\begin{tabular}{cccccccccccc}
				\toprule[1pt]
				{Datasets}&{Quantiles}&GNUM-PL&GNUM-GCN&GNUM-GAT&GNUM-CT&Two-Model&CTM&Uplift-RF&NetDeconf&DML&DRL\\
				\hline
				\multirow{2}*\textbf{Industry-A}  & {Top10\%}  &   \textbf{6.92\%}  & 6.58\%  & \underline{6.79\%} & 6.67\% & 4.18\%  &5.50\% & 6.00\% & 6.54\% & 6.50\% & 6.44\%  \\
				~ & {Top20\%}  &   \textbf{5.52\%}  & 5.22\%  & \underline{5.45\%} & 5.34\% & 3.12\% & 3.63\% & 4.19\% & 5.19\% & 5.02\% & 4.86\% \\
				\multirow{2}*\textbf{Industry-B}  & {Top10\%}  &  \textbf{8.72\%} & 8.38\%  & \underline{8.51\%} & 8.47\% & 6.44\%   & 6.92\% & 7.51\% & 8.35\% & 8.21\% & 7.98\% \\
				~ & {Top20\%}  & \textbf{8.20\%} & 7.85\% & \underline{8.11\%}  & 7.97\%  & 5.95\%  & 6.21\% & 6.58\%  & 7.74\% & 7.67\%  & 7.32\%\\
				\bottomrule[1pt]
		\end{tabular}}
	\end{center}
\end{table*}

\subsection{Overall Performance}

\subsubsection{Results on BlogCatalog}
\label{sec:blogcatalog}

Since we have both factual and counterfactual outcomes for each user in BlogCatalog, we can measure the performance of different methods by comparing the predicted average treatment effect (ATE) with the ground-truth ATE, where ATE is defined as the average uplift over the users as $ATE=\frac{1}{n} \sum_{i=1}^n \tau_i$. Specifically, we use two evaluation metrics, i.e. the Rooted Precision in Estimation of Heterogeneous Effect ($\epsilon_{P E H E}=\sqrt{\frac{1}{n} \sum_{i=1}\left(\hat{\tau}_i-\tau_i\right)^2}$) and Mean Absolute Error on ATE ($\epsilon_{A T E}=\left|\frac{1}{n} \sum_{i=1}\left(\hat{\tau}_i\right)-\frac{1}{n} \sum_{i=1}\left(\tau_i\right)\right|$). Furthermore, we also extend the setting to the scenario of the discrete outcome, by setting the outcome of each sample greater than the mean ATE value as $1$, otherwise as $0$. It is worth noting that these two metrics require both factual and counterfactual outcomes. Therefore, we only report the results on semi-synthetic dataset BlogCatalog in Table \ref{tab:ate}. Then we have the following observations:
\begin{itemize}
    \item In the regression setting, our proposed method GNUM-CT outperforms other baseline methods by $5$\% to $10$\%, which demonstrates that the proposed class-transformed target is able to utilize the graph-based data more effectively. 
    \item In the classification setting, our proposed method GNUM-PL outperforms GNUM-CT and other baseline methods by $12$\% to $25$\%, which demonstrates that the proposed partial-label learning can further utilize the labeled graph data to improve the overall performance. 
    \item All the graph-based methods achieve a substantial gain over non-graph-based methods, which demonstrates the importance of graph data for uplift modeling. 
    \item The result that the Two-Model achieves bad results demonstrates that it is very important to use a common uplift estimator to utilize the treatment and control group of data together. 
\end{itemize}

\subsubsection{Results on Industrial Datasets}
\label{sec:uplift_curve}
Without the counterfactual outcome, we use commonly accepted metrics, i.e. uplift curve and the Qini Coefficient value to evaluate the performance of different methods on our industrial datasets. 

In detail, we sort all users in the treatment group and control group based on their predicted uplift values. Then we will select the top-$k$\% users to get their uplift $Y_k$ as: 

\begin{equation} 
	Y_k=\sum\limits_{i \in V^k_t}{\hat{\tau}_i}/|V^k_t| - \sum\limits_{i \in V^k_c}{\hat{\tau}_i}/|V^k_c|
\label{eqn:exp_uplift}
\end{equation}
where $V^k_t$ and $V^k_c$ are the set of the top $k$\% samples of the treatment and control group, and $\hat{\tau}_i$ is the predicted uplift value by the model. By changing the $k$ from $10$\% to $100$\%, we can obtain the curve on $Y_k$. Note that the leftmost point corresponds to the top $10$\% users which the models predict as the most sensitive to the treatment and the following right part corresponds to the  top $20$\% users. Since the ATE of the dataset is fixed, all the curves will converge to the same value. A well-performing model will show a curve with a larger slope. The left part has larger values than other methods. The results of uplift curve is shown in Figure \ref{fig:cum_uplift}(a)(b). In addition, we introduce the Qini metric to measure the overall performance of uplift methods \cite{radcliffe2007using}. Similar to the AUC value, the Qini metric measures the distance between the Qini curve and the random curve. The detail of the calculation process of Qini curve and Qini Coefficient can be referred in \cite{radcliffe2007using}. We show the results of the Qini Coefficient in Figure \ref{fig:cum_uplift}(c)(d). 

From Figure \ref{fig:cum_uplift}, we find that the proposed method GNUM-PL and GNUM-CT consistently outperform the baseline methods on two metrics. Specifically, GNUM-PL improves the best performing baseline method NetDeconf by an improvement of $21$\% and $14$\% on two datasets in terms of Qini Coefficient. It demonstrates that our proposed methods have a better ranking performance regarding the user's uplift. Additionally, the result that GNUM-PL performs better than GNUM-CT demonstrates that in the classification scenario, the proposed uplift estimators based on partial label learning can further improve the overall performance because partial label learning can utilize the labeled data more effectively. The result that CTM outperforms Two-Model demonstrates that a common uplift estimator to model both the treatment and control group data is very essential. 

In many real-world scenarios, we only focus on samples with uplift values ranking ahead because we will only give actions to users with larger uplift values. Therefore, we give a detailed comparison of the users with the largest $20$\% uplift values in Table \ref{tab:up}. We can find that GNUM-CT and GNUM-PL also achieve better results than other baseline methods, which demonstrates that our proposed methods with the two estimators can find users with larger uplift.

\begin{table*}
	\caption{The mean-squared error of the inferred uplift between different users.}
	\label{tab:ana}
	\begin{center}	
		\small
		\resizebox{\textwidth}{!}{
			\begin{tabular}{cccccccccccc}
				\toprule[1pt]
				{Datasets}&{Sampling strategy   }&GNUM-PL&GNUM-GCN&GNUM-GAT&GNUM-CT&NetDeconf&Two-Model&CTM&Uplift-RF&DML&DRL\\
				\hline
				\multirow{2}*\textbf{Industry-A} & Neighbors & ${1.05*10^{-3}}$ & ${0.97*10^{-3}}$ & ${1.07*10^{-3}}$ & ${1.11*10^{-3}}$ & ${1.07*10^{-3}}$ & ${1.29*10^{-3}}$   &${1.30*10^{-3}}$  & ${1.22*10^{-3}}$ &  ${1.14*10^{-3}}$ & ${1.18*10^{-3}}$  \\
				~ & Random  & ${1.40*10^{-3}}$ & ${1.38*10^{-3}}$& ${1.40*10^{-3}}$ & ${1.39*10^{-3}}$ & ${1.37*10^{-3}}$
				& ${1.35*10^{-3}}$  &${1.41*10^{-3}}$  & ${1.39*10^{-3}}$  & ${1.36*10^{-3}}$ & ${1.38*10^{-3}}$\\
				\multirow{2}*\textbf{Industry-B} & Neighbors  &  ${1.44*10^{-3}}$ & ${1.32*10^{-3}}$  & ${1.42*10^{-3}}$ & ${1.49*10^{-3}}$ & ${1.48*10^{-3}}$ & ${1.75*10^{-3}}$   & ${1.68*10^{-3}}$ & ${1.63*10^{-3}}$  & ${1.50*10^{-3}}$ & ${1.55*10^{-3}}$\\
				~ & Random  & ${1.95*10^{-3}}$ & ${1.94*10^{-3}}$  & ${1.93*10^{-3}}$ & ${1.95*10^{-3}}$ & ${1.91*10^{-3}}$  & ${1.87*10^{-3}}$   & ${1.90*10^{-3}}$ & ${1.88*10^{-3}}$  & ${1.93*10^{-3}}$ & ${1.92*10^{-3}}$ \\
				\bottomrule[1pt]
		\end{tabular}}
	\end{center}
\end{table*}

\begin{figure*}[htb]
	\centering
	\begin{minipage}{0.24\textwidth}  
		\small
		\centerline{\includegraphics[width=1\columnwidth]{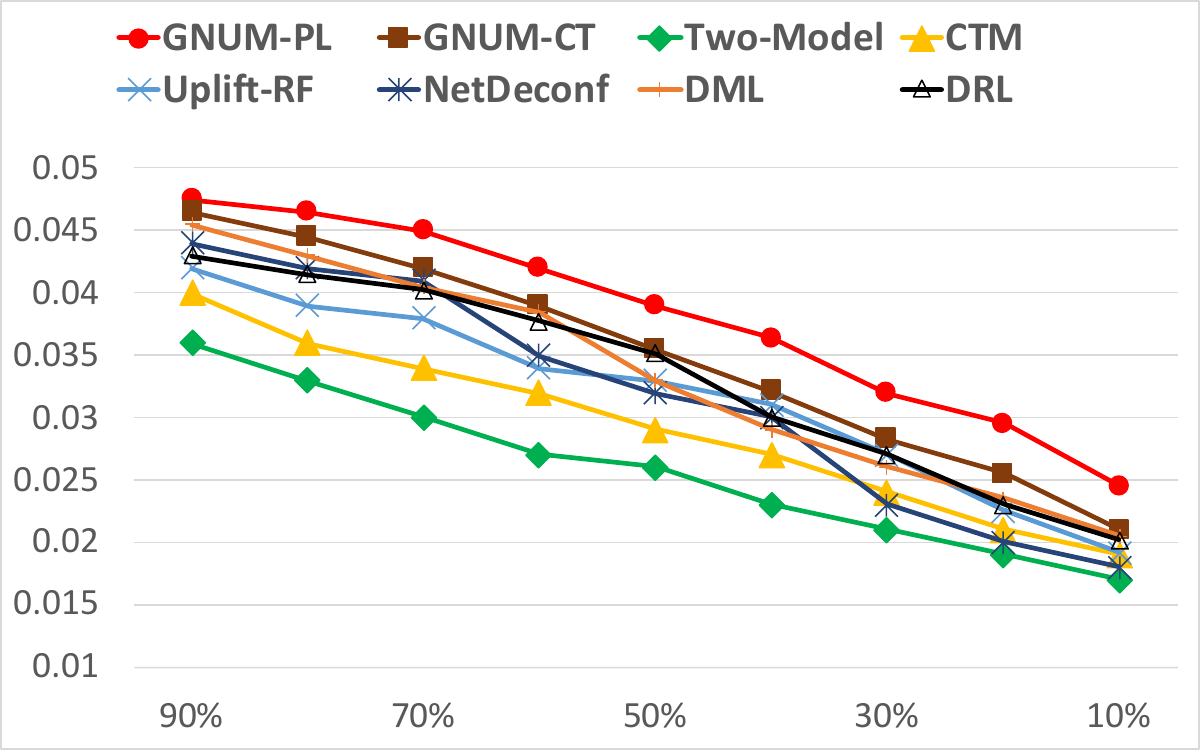}}
		\centerline{(a) Industry-A.}
	\end{minipage}
	\begin{minipage}{0.24\textwidth}
		\small
		\centerline{\includegraphics[width=1\columnwidth]{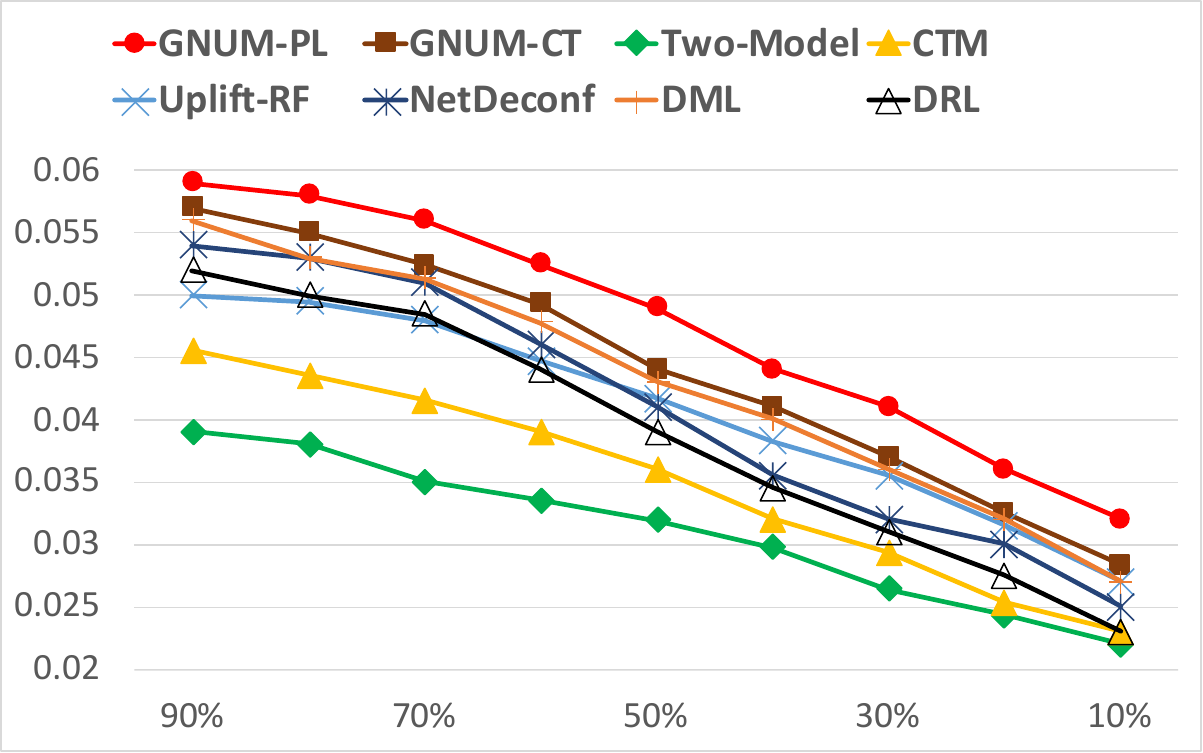}}
		\centerline{(b) Industry-B.}
	\end{minipage}
	\begin{minipage}{0.24\textwidth}  
		\small
		\centerline{\includegraphics[width=1\columnwidth]{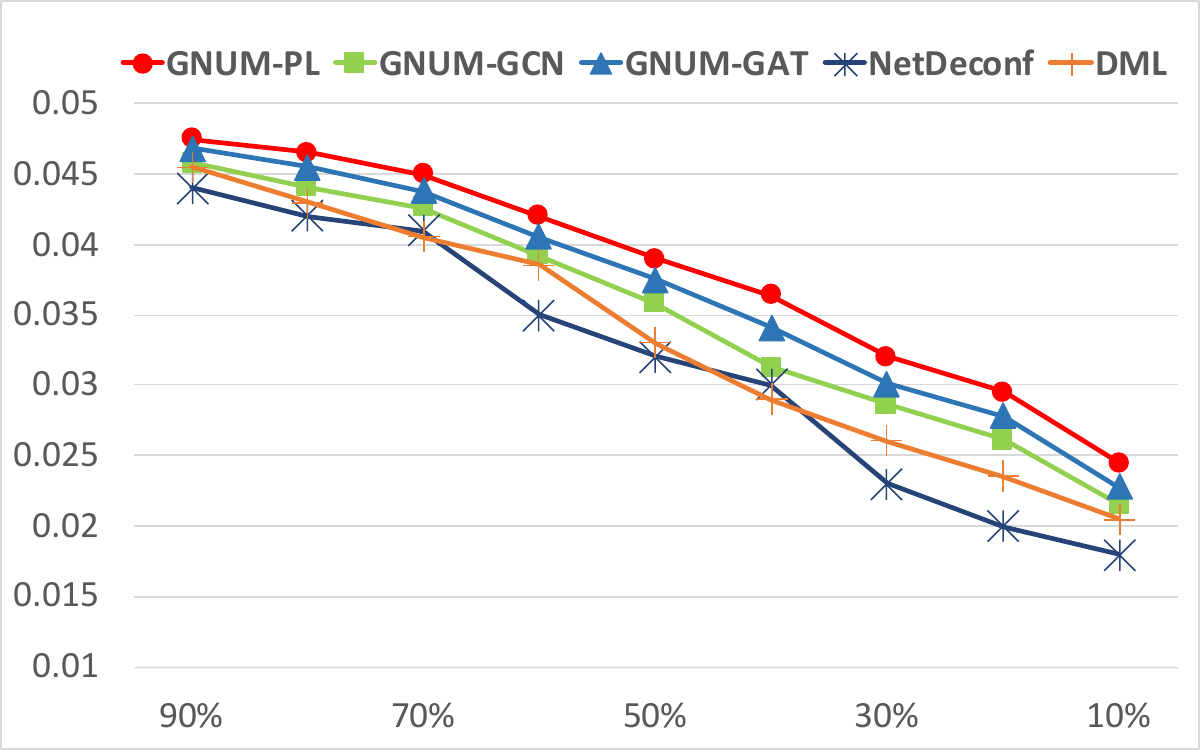}}
		\centerline{(c) Industry-A.}
	\end{minipage}
	\begin{minipage}{0.24\textwidth}
		\small
		\centerline{\includegraphics[width=1\columnwidth]{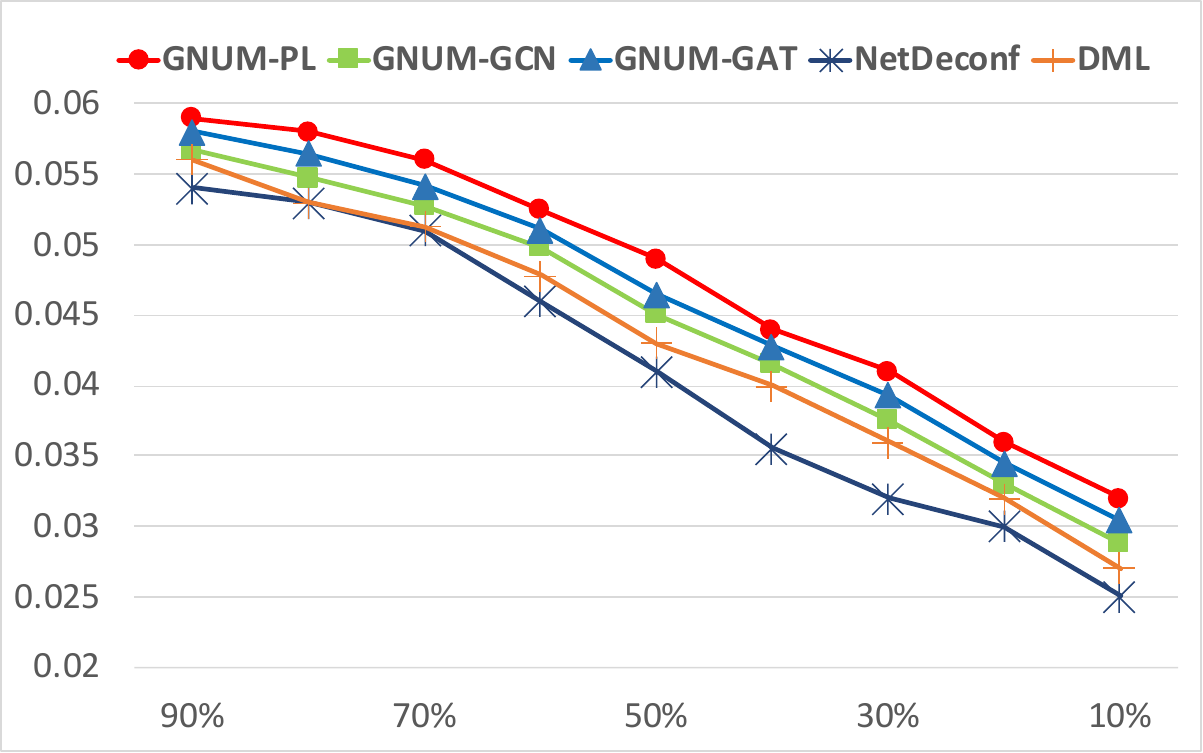}}
		\centerline{(d) Industry-B.}
	\end{minipage}
	\caption{The Qini Coefficient of the different uplift methods under different percentage of labeled users. The ordinate represents the value of Qini Coefficient, and the abscissa represents different percentage of labeled users. (c)(d) More comparisons by replacing the GNN layers in GNUM-PL with GCN/GAT.  }
	\label{fig:cum_rob}	
\end{figure*}

\subsection{In-Depth Analysis}
\subsubsection{Generality to different backbones of GNN models}
\label{sec:gene_GNN}
\begin{figure}[htb]
	\centering
	\begin{minipage}{0.23\textwidth}  
		\small
		\centerline{\includegraphics[width=1\columnwidth]{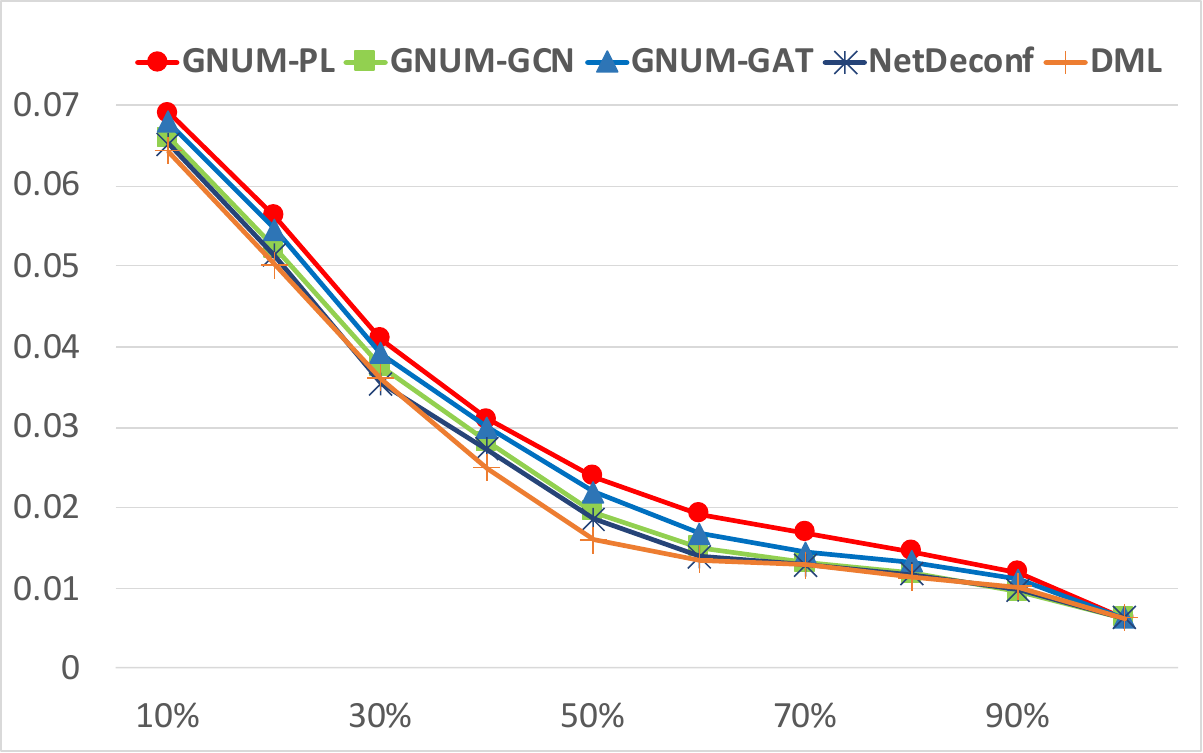}}
		\centerline{(a) Industry-A.}
	\end{minipage}
	\begin{minipage}{0.23\textwidth}
		\small
		\centerline{\includegraphics[width=1\columnwidth]{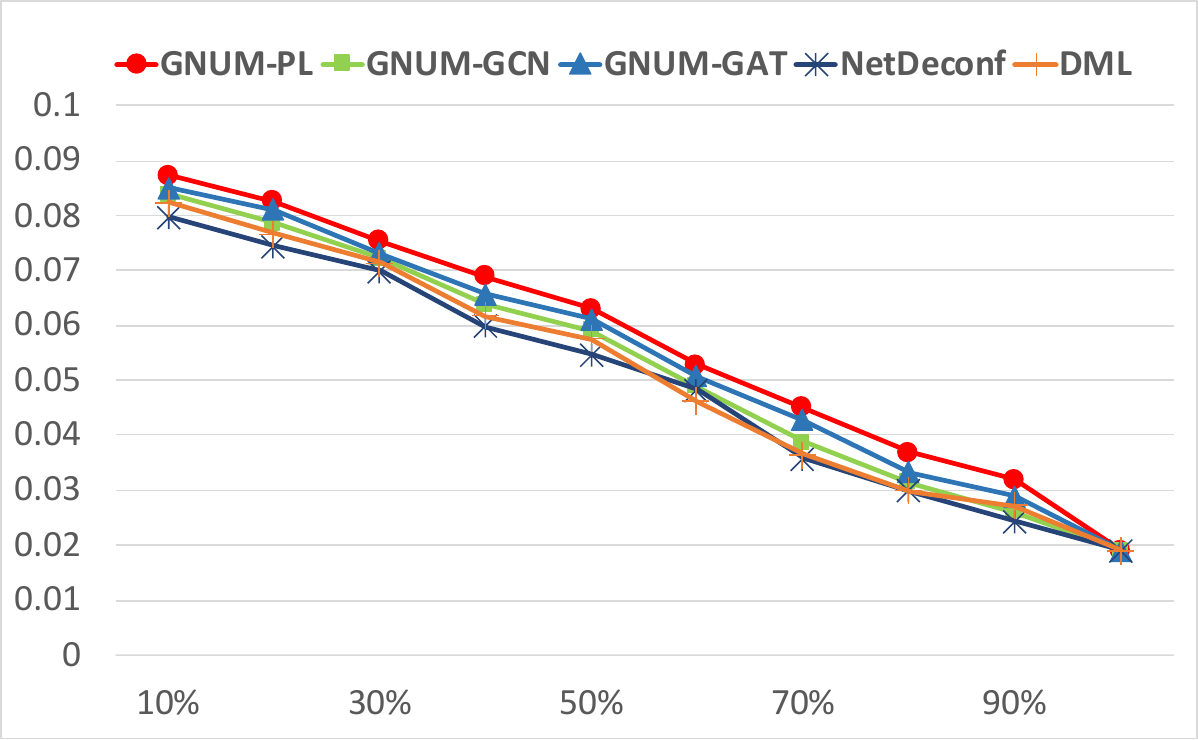}}
		\centerline{(b) Industry-B.}
	\end{minipage}
	\caption{The uplift curve of different graph-based methods. }
	\label{fig:cum_uplift2}	
\end{figure}

We further replace our GNN models with GAT and GCN to demonstrate the generality of our proposed uplift estimators. The results are shown in Figure \ref{fig:cum_uplift2}. We find in most cases, our proposed graph-based methods GNUM-PL, GNUM-GCN and GNUM-GAT perform better than NetDeconf and DML. It demonstrates that our proposed uplift estimators can be adapted to different GNN-based representation learning methods effectively. Furthermore, GNUM-PL still achieves the best performance because our breadth and depth aggregators can utilize more informative information from the social graph. 

\subsubsection{Correlation between Uplift and Social Relationship}
\label{sec:neigh}
Previously, we claim that the uplift difference between users with social relationships is smaller than the uplift difference between random users. To verify this, for each user we first calculate the average of the inferred uplift value of his neighbors. Then, according to the number of his neighbors, we randomly sample the same number of users to calculate the average of their inferred uplift values. Finally, we compare the uplift difference between the above two averaged values and the user’s own inferred uplift value. Specifically, we use mean-squared error (MSE) as the evaluation metric.

The result of the uplift analysis is shown in Table \ref{tab:ana}.
We can see that for the graph-based methods, the uplift difference between neighbors is significantly smaller than the difference between random users compared with non-graph-based methods. 
Compared with the random sampling strategy, the MSE of inferred uplift calculated from neighbors drops by more than 30\%.
It demonstrates that the graph-based method can indeed learn the similarity information from neighbors which is useful for uplift modeling. It is worth mentioning that the smaller difference of uplift is not representing that the model performs better necessarily. The result of GNUM-GCN is worse than that of GNUM-PL and GNUM-GAT because GNUM-GCN does not capture the attentional weight of different neighbors.
The uplift differences between the neighbors of Two-Model and CTM are largest, since they can not capture structural information at all. Therefore, this result demonstrates that it is very essential to utilize social relations to do the uplift estimation. And the performance gain can be achieved by proposing the graph-based model to mine the social relationships between users. 

\subsubsection{Data Scarcity Analysis}
\label{sec:scarcity}
Due to the labeled data scarcity problem for uplift modeling, we propose GNUM-PL and GNUM-CT in this paper to alleviate the problem. To prove this point, we randomly sample $10\%$ to $90\%$  of the labeled users to train different uplift models and compare their performance on the same test set. Note that we will not sample the edges of the graph. We use the Qini Coefficient to evaluate the performance and note that the uplift curve is consistent with the Qini coefficient. The experimental results are shown in Figure ~\ref{fig:cum_rob}(a)(b) and we have the following observations: 
\begin{itemize}
    \item The performance of GNUM-PL and GNUM-CT are consistently above the curves of baseline methods, which demonstrates that both the partial-label-based uplift estimator and the class-transformed uplift estimator are robust to the label scarcity. Furthermore, in the classification setting, the partial-label-based uplift estimator can further improve the performance by utilizing more labeled data explicitly. 
    \item Comparing Netdeconf and other baseline methods, we find that NetDeconf drops more quickly than other baseline methods, which proves our assumption that data scarcity is a more severe problem for the graph-based uplift method because of the more parameters. Therefore, our solution for alleviating the labeled data scarcity problem is very critical. 
\end{itemize}

Moreover, as is shown in Figure ~\ref{fig:cum_rob}(c)(d), GNUM-GCN and GNUM-GAT are still robust to the data scarcity problem. It demonstrates that our partial-label-based uplift estimator can alleviate the data scarcity problem when using different types of GNN-based models. 

\section{Conclusion}
In this paper, we propose GNUM, a novel and general GNN-based framework with two uplift estimators for user uplift modeling. The first uplift estimators are general to different uplift scenarios, which can utilize the two groups of data together to estimate the user uplift. Specifically, when the outcome is discrete, we further propose an uplift estimator based on our designed partial labels to address the labeled data scarcity problem. Experimental results demonstrate that our proposed methods outperform state-of-the-art uplift methods under various evaluation metrics.  We also give an analysis of the relationship between the uplift and social relations. And we further demonstrate the robustness of our proposed model when labeled data is limited.  In the future, we expect to utilize more types of graphs to estimate the user uplift. 

\section{Acknowledgement}
Kun Kuang's research was supported by National Natural Science Foundation of China (62006207), Young Elite Scientists Sponsorship Program by CAST (2021QNRC001), and the Fundamental Research Funds for the Central Universities (226-2022-00142)

\bibliographystyle{unsrt}
\bibliography{gnum}
\clearpage

\appendix
\section{Appendix}
\subsection{Proposition Proof}
\label{appx:proof}

\begin{proof}
\begin{equation} 
\begin{split}
    &\mathbb{E}(z_i|\mathbf{x}_i,\mathit{G})=\mathbb{E}(y_i^{obs}\cdot \frac{t_i-p}{p(1-p)}|\mathbf{x}_i,\mathit{G})  \\ 
    &=\mathbb{E}(t_iy_i^{obs}\cdot \frac{t_i-p}{p(1-p)}+(1-t_i)y_i^{obs}\cdot \frac{t_i-p}{p(1-p)}|\mathbf{x}_i,\mathit{G})
\end{split}
\label{eqn:proof1}
\end{equation}
Given $t_i\in\{0,1\}$, we have $t_i^2=t_i$ and $(1-t_i)^2=(1-t_i)$. Simultaneously with the definition in Eq. \eqref{eqn:y_obs}, we have:
\begin{equation}
\left\{
             \begin{array}{lr}
             t_iy_i^{obs}=t_i^2y_i(1) + t_i(1 - t_i)y_i(0)=t_i^2 y_i(1)=t_iy_i(1) \nonumber \\ 
(1-t_i)y_i^{obs}=t_i(1-t_i)y_i(1) + (1 - t_i)^2y_i(0)=(1-t_i)y_i(0) . \nonumber
             \end{array}
\right.
\end{equation}

Then we can rewire Eq. \ref{eqn:proof1} as \footnote{Due to the limit of space, $t_i^j$ represents $t_i=j$ here.}: 
\begin{equation} 
\begin{split}
    &\mathbb{E}(t_iy_i^{obs}\cdot \frac{t_i-p}{p(1-p)}+(1-t_i)y_i^{obs}\cdot \frac{t_i-p}{p(1-p)}|\mathbf{x}_i,\mathit{G}) \\
    &=\mathbb{E}(t_iy_i(1)\frac{t_i-p}{p(1-p)}+(1-t_i)y_i(0) \frac{t_i-p}{p(1-p)}|\mathbf{x}_i,\mathit{G},t_i^1)*p(t_i^1|\mathbf{x}_i,\mathit{G})\\
    &+\mathbb{E}(t_iy_i(1)\frac{t_i-p}{p(1-p)}+(1-t_i)y_i(0) \frac{t_i-p}{p(1-p)}|\mathbf{x}_i,\mathit{G},t_i^0)*p(t_i^0|\mathbf{x}_i,\mathit{G}) \\
    &=\mathbb{E}(\frac{y_i(1)}{p}|\mathbf{x}_i,\mathit{G})*p(t_i=1|\mathbf{x}_i,\mathit{G})-\mathbb{E}(\frac{y_i(0)}{1-p}|\mathbf{x}_i,\mathit{G})*p(t_i=0|\mathbf{x}_i,\mathit{G})\\
    &=\mathbb{E}(y_i(1)|\mathbf{x}_i,\mathit{G})\frac{p(t_i=1|\mathbf{x}_i,\mathit{G})}{p}-\mathbb{E}(y_i(0)|\mathbf{x}_i,\mathit{G})\frac{p(t_i=0|\mathbf{x}_i,\mathit{G})}{1-p}\\
    &=\mathbb{E}(y_i(1)|\mathbf{x}_i,\mathit{G})-\mathbb{E}(y_i(0)|\mathbf{x}_i,\mathit{G}) \\
    &=\hat{\tau}_i \nonumber
\end{split}
\end{equation}
\end{proof}

\subsection{Pseudo Code and Complexity Analysis}
\label{appx:code}
The pseudo-code of GNUM-CT is given in Algorithm \ref{alg2} and GNUM-PL in Algorithm \ref{alg1}. Our proposed method can be trained by an end-to-end back-propagation, and thus we can use gradient descent to optimize the model.

For each sample, it will go through the GNUM layer first and then perform the uplift prediction. The complexity of the whole process is
$O\left(M\sum_{i=0}^L f_l+N \sum_{i=1}^L f_{i-1}f_i \right)$ where $N$ denotes the total number of users, including the users in the treatment and control group, $M = \left| \mathit{E}\right|$ is the number of relationships, $L$ is number of hidden layers and $f_l$ is the dimensionality of the $l^{th}$ hidden layer. In practice, $L$ and $f_l$ are often bounded within a small constant. The proposed method is linear to the number of users and number of relationships in the dataset respectively. Therefore, the overall model is scalable. 

$O(Nd^2_{max}S_{max})$, where $N$ denotes the total number of users, including the users in the treatment and control group, $d_{max}$ denotes the maximum number of dimensionality among different layers and $S_{max}$ denotes in the social graph the maximum degree among all the users. In practice, $S_{max}$ is often bounded within a constant. For example, in social networks like Facebook, the number of friends a user can have has an upper bound. 

\begin{algorithm}[t]\caption{Graph Neural Network for Uplift Modeling (GNUM) using Class-Transformed Target} \label{alg2}
	\begin{algorithmic}[1]
		\REQUIRE Set of user features, treatment assignment and observed outcome $\{(\mathbf{x}_i,t_i,y_i)\}_{i=1}^N$, social graph $\mathit{G}=(\mathit{V},\mathit{E})$ and number of layers $L$, 
		\ENSURE Prediction of user uplift and GNUM parameters $\mathbf{\Theta}$.
		\STATE Initialize all parameters $\mathbf{\Theta}$ and using $\mathbf{X}$ as $\mathbf{H}^{(0)}$
		\WHILE {$\mathcal{L}$ does not converge}
		\FOR{i $\leftarrow$ 1 to L}
		\STATE Calculate graph-based representation $\mathbf{H}^{(l)}$.
		\ENDFOR
		\STATE Calculate  $\hat{y}_i^{pl1}$ and $\hat{y}_i^{pl2}$  from learned representations $\mathbf{H}^{(L)}$ using Eq. \eqref{eqn:partial_resp}.
		\STATE Calculate the class-transformed target $z_i$ using Eq. \eqref{eq_zi}.
		\STATE Calculate the loss $\mathcal{L}_{CT}$ using Eq. \eqref{eq:loss_ct}.
		\STATE Update $\mathbf{\Theta}$ using back-propagation.
		\ENDWHILE
		\STATE Using the trained model parameters to estimate the user uplift as: $\hat{\tau}_i=\mathbb{E}(z_i|\mathbf{x}_i, \mathit{G})$.
	\end{algorithmic}
\end{algorithm}

\begin{algorithm}[t]\caption{Graph Neural Network for Uplift Modeling (GNUM) using Partial Label Learning} \label{alg1}
	\begin{algorithmic}[1]
		\REQUIRE Set of user features, treatment assignment and observed outcome $\{(\mathbf{x}_i,t_i,y_i)\}_{i=1}^N$, social graph $\mathit{G}=(\mathit{V},\mathit{E})$ and number of layers $L$, 
		\ENSURE Prediction of user uplift and GNUM parameters $\mathbf{\Theta}$.
		\STATE Initialize all parameters $\mathbf{\Theta}$ and using $\mathbf{X}$ as $\mathbf{H}^{(0)}$
		\WHILE {$\mathcal{L}$ does not converge}
		\FOR{i $\leftarrow$ 1 to L}
		\STATE Calculate graph-based representation $\mathbf{H}^{(l)}$.
		\ENDFOR
		\STATE Calculate  $\hat{y}_i^{pl1}$ and $\hat{y}_i^{pl2}$  from learned representations $\mathbf{H}^{(L)}$ using Eq. \eqref{eqn:partial_resp}.
		\STATE Construct the sample's partial labels based on $y_i$ and $t_i$.
		\STATE Calculate the loss $\mathcal{L}$ using Eq. \eqref{eq:gnum_loss}.
		\STATE Update $\mathbf{\Theta}$ using back-propagation.
		\ENDWHILE
		\STATE Using the trained model parameters to estimate the user uplift as: $\hat{\tau}_i = 1 - P(S_i=[1, 0, 0]|\mathbf{x}_i,\mathit{G}) - P(S_i=[0, 0, 1]|\mathbf{x}_i,\mathit{G})$
	\end{algorithmic}
\end{algorithm}

\subsection{More Experimental Details}
\subsubsection{More Details about Dataset}
\label{appn:dataset}
Blogcatalog is a social blog directory that manages bloggers and their blogs. The dateset contains 
the features of bloggers and the social relationships between bloggers listed on the BlogCatalog website. The features are bag-of-words representations of keywords in bloggers’ descriptions. 
We follow the assumptions and procedures
of synthesizing the outcomes and treatments assignments in \cite{guo2020learning}, in which the outcomes represent the opinions of readers on each
blogger and the treatments represent whether contents created by a blogger receive more views on mobile devices or desktops.
In the treatment group, the blogger’s blogs are read more on mobile devices. In the control group, the blogger’s blogs are read more on desktops. 
We assume that the social relationships of bloggers can causally affect their treatment assignments and readers’ opinions of them. 
We trained the LDA topic model to get the device
preference of the readers of the i-th blogger’s content as:

\begin{center}
$
\begin{aligned}
\operatorname{Pr}(t&\left.=1 \mid \mathbf{x}_i, \mathrm{~A}\right)=\frac{\exp \left(p_1^i\right)}{\exp \left(p_1^i\right)+\exp \left(p_0^i\right)} ; \\
p_1^i &=\kappa_1 r\left(\mathbf{x}_i\right)^T r_1^c+\kappa_2 \sum_{j \in \mathcal{N}(i)} r\left(\mathbf{x}_j\right)^T r_1^c \\
&=\kappa_1 r\left(\mathbf{x}_i\right)^T r_1^c+\kappa_2\left(\mathrm{~A} r\left(\mathbf{x}_j\right)\right)^T r_1^c ; \\
p_0^i &=\kappa_1 r\left(\mathbf{x}_i\right)^T r_0^c+\kappa_2 \sum_{j \in \mathcal{N}(i)} r\left(\mathbf{x}_j\right)^T r_0^c \\
&=\kappa_1 r\left(\mathbf{x}_i\right)^T r_0^c+\kappa_2\left(\mathrm{~A} r\left(\mathbf{x}_j\right)\right)^T r_0^c,
\end{aligned}
$
\end{center}

where $\kappa_1, \kappa_2 \geq 0$ signifies the magnitude of the confounding bias resulting from a blogger’s topics and her neighbors’ topics, respectively.  $\kappa_1 = 0, \kappa_2 = 0$ means
the treatment assignment is random and there is no selection bias,
and greater $\kappa_1, \kappa_2$ means larger selection bias. The factual outcome and the counterfactual outcome of the i-th blogger are simulated as:

\begin{center}
$
\begin{aligned}
y^F\left(\mathbf{x}_i\right) &=y_i=C\left(p_0^i+t_i p_1^i\right)+\epsilon \\
y^{C F}\left(\mathbf{x}_i\right) &=C\left[p_0^i+\left(1-t_i\right) p_1^i\right]+\epsilon,
\end{aligned}
$
\end{center}
where $C$ is a scaling factor and the noise is sampled as $\epsilon \sim \mathcal{N}(0,1)$. In this work, we set $C=5, \kappa_1=10, \kappa_2 \in\{0.5,2\}$.

\subsubsection{Details about the comparing methods}
\label{appn:baseline}
The detail descriptions of the comparing methods can be found here. 
\begin{itemize}
	\item{Two-Model: This method infers the labels in the treatment group and the control group respectively. }
	\item{CTM: The Class-Transformation model (CTM) creates the new target variable to estimate the uplift.}
	\item{Uplift-RF}: This model modifies existing random forest algorithms to directly infer the uplift.
	\item{DML}: Double Machine Learning (DML) method uses the Neyman orthogonal score and cross-fitting to construct the uplift estimator.
	\item{DRL}: Doubly Robust Learning (DRL) method combines the error imputation and the inverse propensity score estimator to address the bias problem.
	\item{NetDeconf}: It uses the graph to minimize confounding bias in the task of estimating treatment effects.
	\item{GNUM-GCN/GAT}: To show the effectiveness of partial label learning, we replace the graph representation learning method with GCN and GAT respectively for comparison.
\end{itemize}

\subsubsection{Parameter Settings}
\label{appn:para}
We adopt the deep neural networks \cite{liu2017survey} with two $64$-unit hidden layers as the building block for the Two-Model and CTM methods. For uplift random forest, we set the number of estimators as $50$ and the depth of the tree as $5$. For NetDeconf and our proposed method, we use two layers of GNN with units of $128-64$. All the weight matrices are initialized using Xavier initialization \cite{glorot2010understanding}.  We train the model for $5$ epochs with a learning rate of $0.0001$ and batch size of $256$. Note that we use grid search to get the best hyper-parameters. For each dataset, we randomly sampled 70\% of the users as the training set, 10\% as the validation set, and 20\% as the test set.  The models are trained on a cluster of 10 Dual-CPU servers with AGL \cite{zhang2020agl} framework. For the largest dataset, containing 573.8K nodes and 2.6M edges, the proposed method took about 17 minutes to train on a cluster of 10 Dual-CPU servers. We run each algorithm $5$ times and report the average result.

\end{document}